\documentclass[letterpaper, 10 pt, conference]{ieeeconf}  

\IEEEoverridecommandlockouts                              

\usepackage{graphicx}

\newtheorem{definition}{Definition}
\usepackage{amsmath}
\usepackage{amsfonts}
\usepackage{amssymb}
\usepackage{algorithm}
\usepackage{algpseudocode}

\title{\LARGE \bf
Distributionally Safe Reinforcement Learning under Model Uncertainty: A Single-Level Approach by Differentiable Convex Programming}

\author{Alaa Eddine Chriat$^{1}$ and Chuangchuang Sun$^{1}$
\thanks{$^{1}$The authors are with the Aerospace Engineering Department, Mississippi State University, Starkville, MS 39759, USA. Emails:
        {\tt\small aec652@msstate.edu, csun@ae.msstate.edu}.}%
}

\usepackage{xcolor}

\graphicspath{{figures/}}

\usepackage{amsfonts}
\usepackage{microtype}
\usepackage{graphicx}
\usepackage{subfigure}
\usepackage{booktabs} 
\usepackage{balance}
\usepackage{makecell}
\usepackage{bm}
\makeatletter
\let\NAT@parse\undefined
\makeatother
\usepackage{hyperref}
\usepackage[numbers,sort&compress]{natbib}

\let\labelindent\relax

\usepackage{amssymb}
\usepackage{multirow}
\usepackage{amsmath}
\usepackage{algorithm,algpseudocode}
\usepackage{wrapfig}

\newcommand{\bea}{\begin{eqnarray}}
\newcommand{\eea}{\end{eqnarray}}
\newcommand{\beas}{\begin{eqnarray*}}
\newcommand{\eeas}{\end{eqnarray*}}
\newcommand{\leftm}{\left[\begin{array}}
\newcommand{\rightm}{\end{array}\right]}

\usepackage{diagbox,hhline}
\usepackage{enumitem}

\def\span{{\mbox{\bf span}}}

\def\proj{{\mbox{\text{Proj}}}}

\definecolor{commentclr}{RGB}{110, 149, 204}

\usepackage{stackengine}
\stackMath

\usepackage{color}
\usepackage{soul}
\usepackage{xspace}
\usepackage{cleveref}
\usepackage{makecell}
\usepackage{balance}

\begin{document}

\maketitle
\thispagestyle{empty}
\pagestyle{empty}

\begin{abstract}
Safety assurance is uncompromisable for safety-critical environments with the presence of drastic model uncertainties (e.g., distributional shift), especially with humans in the loop. 
However, incorporating uncertainty in safe learning will naturally lead to a bi-level problem, where at the lower level the (worst-case) safety constraint is evaluated within the uncertainty ambiguity set. In this paper, we present a tractable distributionally safe reinforcement learning framework to enforce safety under a distributional shift measured by a Wasserstein metric. To improve the tractability, we first use duality theory to transform the lower-level optimization from infinite-dimensional probability space where distributional shift is measured, to a finite-dimensional parametric space. Moreover, by differentiable convex programming, the bi-level safe learning problem is further reduced to a single-level one with two sequential computationally efficient modules: a convex quadratic program to guarantee safety followed by a projected gradient ascent to simultaneously find the worst-case uncertainty. This end-to-end differentiable framework with safety constraints, to the best of our knowledge, is the first tractable single-level solution to address distributional safety. 
We test our approach on first and second-order systems with varying complexities and compare our results with the uncertainty-agnostic policies, where our approach demonstrates a significant improvement on safety guarantees.

%

\end{abstract}

\section{INTRODUCTION}\label{sec:intro}
In many real-world applications, there often can be unmodelled dynamics and uncertainties, both internally (e.g., systems failure/  dysfunctionality, perceptional noise) and externally (e.g., gust, tough terrain). Those uncertainties make control problems and training processes more challenging and susceptible to errors, potentially leading to undesirable/ disastrous outcomes. Moreover, there are often more restrictive settings where environmental configurations change more drastically. For example, there can always be ``unknown unknowns'' during the deployment of autonomous systems, such as off-road autonomy, advanced air mobility, etc.
Also, one of the fundamental challenges towards (super) human-level intelligence for robots is the ability of trained policies to generalize beyond the specific environments they initially ever encountered during training.
However, with the presence of those pervasive uncertainties, safety is not compromisable in many safety-critical scenarios with human health/ lives at stake. For example, in self-driving cars, decision-making in real-time in complex and uncertain environments can be very challenging, considering various surroundings, moving pedestrians, etc.

Learning-based control has attracted lots of attention to combine the advantages of machine learning and modern control theory. On one hand, data-driven machine learning methods have arisen to learn control policies from interactions with the environments without an accurate prior model. On the other hand, while control theory has been extensively investigated and in general has rigorous formal guarantees of performance, it often needs relatively accurate models, which are expensive, if not impossible to obtain. 
To achieve robust and safe decision-making and control with the presence of uncertainties, there has been a large volume of learning-based approaches in the literature. In general, robust learning and control treats uncertainty with a soft criterion and aims to get the performant policy under the (worst-case) uncertainty, yielding a minimax optimization problem~\cite{li2019robust, zhang2019policy, sun2021romax, yang2022stackelberg, zhou2021decentralized, lauffer2022no, bai2021sample}. While they can provide some degrees of safety, it is not guaranteed. On the other hand, in safe learning, safety is imposed as a hard constraint~\cite{jin2021safe, liu2020ipo, achiam2017constrained, bertsekas2015parallel, geibel2005risk, chow2017risk, chow2018lyapunov, stooke2020responsive}. However, incorporating uncertainty in safe learning will naturally lead to a bi-level problem, where at the lower level the (worst-case) safety constraint is evaluated within the uncertainty ambiguity set.

In this paper, we consider a distributinoally safe reinforcement learning (DSRL) problem with a distributional shift measured by a Wasserstein metric. Such distributional shift quantifies drastic uncertainties and it leads to a distributional optimization at the lower level of the overall safe learning problem to evaluate the worst-case safety constraint. Both the distributional optimization and the bi-level nature make the distributional safe learning problem intractable. We first use duality theory to transform the lower-level optimization problem into the finite-dimensional parametric space instead of infinite-dimensional probability space. Moreover, we further reduce the bi-level safe learning problem as single-level by differentiable convex programming~\cite{amos2017optnet} to sequentially find the worst case uncertainty and simultaneously guarantee that the safety constraints are satisfied. Both modules are computationally efficient, with the former as a projected gradient ascent and the latter as a convex quadratic programming. Moreover, this constrained learning pipeline is end-to-end differentiable. To the best of our knowledge, this is the first attempt to address distributional safety with a tractable single-level solution. 


The rest of the paper is organized as follows.
Section \ref{sect:prelim} reviews the preliminaries needed to complete this work. In section \ref{sect: bilevel}, we formulate our distributionally safe approach using distributional robust optimization. In section \ref{sect: reduce} we reduce the problem into a single-level learning-based approach. Section \ref{sect: numerical} contains numerical simulations and results comparisons, and section \ref{sect: conc} has concluding notes of our work.

\subsection{Related works on safe/robust learning} 
There are multiple ways to mitigate quantified uncertainty in the context of safe/ robust learning, such as~\cite{zhang2020robust1, zhang2020robust}, those focusing on the area of robotics~\cite{brunke2021safe, singh2021reinforcement}, and comprehensive surveys~\cite{garcia2015comprehensive, moos2022robust}. Specifically, the uncertainty variable can be treated as a context variable representing different tasks and can be subsequently solved as multi-task or meta-learning problems~\cite{eghbal2021learning, rakelly2019efficient}. Moreover, given optimization theories, robust learning algorithms have also been developed based on interior point methods~\cite{jin2021safe, liu2020ipo}, successive convexification~\cite{achiam2017constrained} and (augmented) Lagrangian methods~\cite{bertsekas2015parallel, geibel2005risk, chow2017risk, chow2018lyapunov, stooke2020responsive}. In learning-based control, Lyapunov theory, model predictive control, and control barrier functions are also employed to develop robust learning algorithms~\cite{choi2020reinforcement, zheng2021safe, cheng2019end, ames2016control, berkenkamp2017safe, sun2021fisar}. Additionally, if the uncertainty is considered in the worst-case scenario, minimax policy optimization~\cite{li2019robust, zhang2019policy, sun2021romax} or its generalization Stackelberg games~\cite{yang2022stackelberg, zhou2021decentralized, lauffer2022no, bai2021sample} are often the frameworks to promote resilience. Other works include meta-adaptive nonlinear control integrating learning modules for fast adaptation in unpredictive settings~\cite{shi2021meta, o2022neural}.

In terms of robust learning and control under \textit{distributional shift}, model-based approaches~\cite{van2015distributionally} such as approximate dynamic programming \cite{yang2020wasserstein,hakobyan2021wasserstein} and model predictive control \cite{bahari2022safe,coulson2021distributionally,coppens2021data} have been proposed, either under a Wasserstein metric or chance-constrained criterion. In the model-free regime, one line of work is to generate environments/ tasks with distributional shifts for policy training to achieve robustness~\cite{ren2022distributionally, morrison2020egad,wang2019adversarial}. Moreover, to balance the worst-case (robustness) and average performance, \cite{xu2022group} trains policies over task groups by adding regularization to the worst possible outcomes. In offline RL~\cite{kallus2022doubly}, \cite{shi2022distributionally} proposes a distributionally robust formulation with tabular Markov decision processes with an uncertainty set specified by the Kullback-Leibler (KL) divergence. Those approaches are often based on duality in the distributionally robust optimization theories~\cite{sinha2017certifiable, blanchet2019quantifying, rahimian2019distributionally}.

\section{PRELIMINARIES}\label{sect:prelim}

\subsection{High-order CBF}

In control theory, control barrier functions play a crucial role in ensuring that a dynamic system can accomplish target objectives while ensuring that safety is not compromised, a control barrier function essentially evaluates the system's safety and returns a scalar quantity. Consequently, our objective is to determine a control input that keeps the system within the safety boundaries as determined by the control barrier function. Mathematically, consider the nonlinear control-affine system:
\begin{equation}\label{nonlinearsystem}
\dot{x}(t)=f(x(t))+g(x(t)) u(t)
\end{equation}
where $f$ and $g$ are globally Lipschitz, $x\in\mathbb{R}^n$ and $u\in\mathbb{R}^m$ are the states and control inputs, respectively, constrained in closed sets, with initial condition $x(t_0) = x_0$. The time dependency on $t$ can be omitted for notational simplicity.
\begin{definition}\label{cbfdef}
\cite{ames2016control}$h: \mathbb{R}^{n} \rightarrow \mathbb{R}$ is a barrier function for the set $C=\left\{x \in \mathbb{R}^{n}: h(x) \geqslant 0\right\}$ if $\exists$ an extended class-$\mathcal{K}$ function $\alpha(\bullet)$ such that:
\begin{equation}\label{cbfeqt}
\begin{gathered}
\sup _{u \in U}[L_{f}h(x)+L_{g}h(x) u+\alpha(h(x))] \geqslant 0 \\
\inf_{\text{int}(C)}[\alpha(h(x)) ]\geqslant 0 \text {  \quad  and  \quad  } \lim_{\partial C} \alpha(h(x))=0
\end{gathered}
\end{equation}
\end{definition}
Because not all systems are first-order in inputs, we can use higher-order control barrier functions to constrain higher-order systems.
\begin{definition}
\cite{xiaohigh}For the {nonlinear} system \eqref{nonlinearsystem} with the $m^{th}$ differentiable function $h(x)$ as a constraint, we define a sequence of functions $\psi_{i}$ with $i \in \{1,2,...,m\}$, starting from $\psi_{0}=h(x)$: $\psi_{i}(x, t)=\dot{\psi}_{i-1}(x, t)+\alpha_{i}\left(\psi_{i-1}(x, t)\right)$.
and define $C_{i}(t)$ sequence of safe sets associated with each $\psi_{i}$: $C_{i}(t)=\left\{x \in \mathbb{R}^{n}: \psi_{i-1}(x, t) \geqslant 0\right\}$.
The function $h(x)$ is a high order control barrier function if there exist extended class-$\mathcal{K}$ functions $\alpha_{i}(\bullet)$ such that $\psi_{m}(x, t) \geqslant 0$.
\end{definition}

Control barrier functions offer significant promise in the development of secure dynamic systems, with applications in various robotic and autonomous systems.

\subsection{Reinforcement learning: DDPG}
Reinforcement Learning agents learn to make sequential decisions by interacting with an environment to achieve a certain goal. 
One popular RL approach is Deep Deterministic Policy Gradient (DDPG), an off-policy algorithm that addresses continuous action space systems.
In DDPG, the agent employs two neural network architectures, the actor and the critic. The actor-network represents the agent's policy, mapping the observed state directly to a specific action in the continuous action space $\mu\left(s \mid \theta^\mu\right)$ and aiming to maximize the cumulative reward function as $J(\theta) = \sum_{k=1}^T \gamma^k {R}(s_k, a_k)$, where $\gamma$ and $R$ are the discount factor and rewards, respectively. The critic-network, on the other hand, evaluates the quality of the actions chosen by the actor $Q\left(s, a \mid \theta^Q\right)$. It estimates the expected cumulative reward, known as the Q-value, by taking both the current state and the action as input $\mathcal{L}(\theta) = \mathbb{E}_{s,a,r,s^\prime}\left((y-Q(s,a|\theta^Q)\right)^2$ and $y = R + \gamma Q^{\prime}\left(s, \mu^{\prime}(s \mid \theta^{\mu^{\prime}}) \mid \theta^{Q^{\prime}}\right)$.
These networks work together to guide the agent into achieving the desired goal. 
DDPG employs a replay buffer, which stores a tuple of past experiences $\langle S, A, \mathcal{R}, S' \rangle$, where $S$ is a set of agent states in the environment, $A$ is a set of agent actions, $\mathcal{R}$ is the reward function, and $S'$ is a set of agent next states. During training, the agent samples random batches of experiences from this buffer to update the actor and critic networks. 
DDPG’s efficiency allows training agents to perform accurate and complex tasks in a wide range of real-world scenarios.

\subsection{Differentiable convex programming}

Differentiable convex programming is a powerful technique that allows computation of the gradients of an optimization problem objective function with respect to the parameters of the problem, by taking matrix differentiation of the Karush-Kuhn-Tucker (KKT) conditions. A notable example of a differentiable optimization method is CVXlayers ~\cite{diffcvxoptlayer}, which incorporates differentiable optimization problems in a disciplined fashion. In a broader sense, this methodology can be applied to differentiate through disciplined convex programs by initially mapping them into cone programs ~\cite{diffconeprog}, computing the gradients, and subsequently mapping back to the original problem. It is worth noting that the convex quadratic program (QP) can be differentiated through the KKT conditions \cite{amos2017optnet}, which serve as equivalent conditions for global optimality. According to the KKT conditions, at the optimal solution, the gradient of the Lagrangian function with respect to the program's input and parameters must be zero. Consequently, by taking the partial derivative of the Lagrangian function with respect to the input and extending it through the chain rule to the program's parameters, their gradients can be obtained. For a generalized QP:
\begin{equation}
\begin{aligned}
\min_x\ \  \frac{1}{2} x^T Q x+q^T x , \quad\text { s.t.}\ \  A x=b \quad G x \leq h, \\ 
\end{aligned}
\end{equation}
we can write the Lagrangian of the problem as:
\begin{equation}
L(z, \nu, \lambda)=\frac{1}{2} z^T Q z+q^T z+\nu^T(A z-b)+\lambda^T(G z-h)
\end{equation}
where $\nu$ are the dual variables on the equality constraints and $\lambda \geq 0$ are the dual variables on the inequality constraints.
The KKT conditions for stationarity, primal feasibility, and complementary slackness are:
\begin{equation}
\begin{aligned}
Q z^{\star}+q+A^T \nu^{\star}+G^T \lambda^{\star} & =0 \\
A z^{\star}-b & =0 \\
D\left(\lambda^{\star}\right)\left(G z^{\star}-h\right) & =0
\end{aligned}
\end{equation}
By differentiating these conditions, we can shape the Jacobian of the problem as follows.
\begin{equation}
\left[\begin{array}{l}
d_z \\
d_\lambda \\
d_\nu
\end{array}\right]=-\left[\begin{array}{ccc}
Q & G^T D\left(\lambda^{\star}\right) & A^T \\
G & D\left(G z^{\star}-h\right) & 0 \\
A & 0 & 0
\end{array}\right]^{-1}\left[\begin{array}{c}
\left(\frac{\partial \ell}{\partial z^{\star}}\right)^T \\
0 \\
0
\end{array}\right]
\end{equation}
Using the chain rule, we can get the derivatives of any loss function with respect to any of the parameters in the QP.


\section{Distributionally Safe Reinforcement Learning under Model Uncertainty}

Stochastic optimization solves problems with stochastic variables under a prior distribution. However, in many real-world engineering applications, such a prior distribution is often inaccurate or can change in the deployment. As a result, Distributionally Robust Optimization (DRO) addresses optimization problems with the presence of uncertainty, whose prior probability distribution might shift.
Compared to robust optimization admitting parameters within certain sets or stochastic optimization admitting a distribution of parameters, DRO takes a more conservative formulation to solve a more challenging problem, where the distribution of the parameters shifts from the prior. Then following the minimax formulation, DRO aims to minimize the objective function with a worst-case distributional shift in the ambiguity set. This general strategy provides an adaptable/ resilient solution that exhibits strong performance in different/ unseen testing environments.


\subsection{Distributional safety under model uncertainty using the Wasserstein metric: a bi-level problem} \label{sect: bilevel}

Control barrier functions have found application in maintaining the safety constraints of control systems. In the realm of reinforcement learning, CBFs can guarantee that the actions taken by an agent comply with safety constraints, while maximizing a cumulative reward function. There are several methods to incorporate the CBF in reinforcement learning. For example, CBF can be incorporated in the reward function of the agent. Specifically, rewarding safe actions and penalizing the agent for violating the safety constraints leads to an agent that is biased towards safe actions. Another way of using CBF in RL is to incorporate the CBF module as a safety shield for the RL actions. The CBF will then monitor the actions and rectify any action that violates the safety constraint to the closest safe action \cite{chriat2023optimality}. In the presence of model uncertainty or noise, the conventional CBF approach does not hold the same efficacy. Some work addressed robust CBFs by estimating the dynamics noise and incorporating it directly into the CBF \cite{dacs2023robust}, while some other works have used the conditional value at risk of the constraints to build a safe CBF~\cite{chriat2023CVar}.

In this section, we aim to integrate distributional shift measured by a Wasserstein metric into CBF to guarantee the safety of the system under the model uncertainty. Consider the nonlinear system with additive uncertainty: 
\begin{equation}\label{noisysystem}
\dot{x}(t)=f(x(t))+g(x(t)) u(t) + \omega 
\end{equation}
where $x\in\mathbb{R}^n$ and $u\in\mathbb{R}^m$ and $\omega\sim p_0(\omega)$ are the states, control inputs, and the disturbance acting on the system, respectively. Here $p_0(\omega)$ is the prior probability distribution of the model uncertainty.
The metric to measure the distance between two probability distributions, i.e., distributional shift, is the Wasserstein metric ($W_d(\bullet,\bullet)$), which is often the case in many works addressing distributional robustness~\cite{givens1984class}. Comparing the Wasserstein metric with the KL divergence~\cite{joyce2011kullback}, the most obvious advantage of the former is symmetry (i.e., $W_d(p_1, p_2) = W_d(p_2, p_1)$), which however does not hold for the latter in general. 
Then the ambiguity set of the perturbed distribution $p(\omega)$ from the nominal distribution $p_0(\omega)$ under a Wasserstein metric~\cite{givens1984class} is expressed as $\mathcal{P} = \{p(\omega)\in p(\mathbb{W})|W_d\big(p(\omega), p_0(\omega)\big)\le \rho\}$, with $\rho$ as the threshold of such a shift. Moreover, $\mathbb{W}$ is the support of $p(M)$ and is assumed to be convex and closed, which is a common assumption in related works. 


The objective of safe reinforcement learning is to generate a control input $u_r$ to achieve certain goals characterized by the reward function in the MDP while satisfying safety constraints. The typical way for goal-reaching robotic navigation is to drive a potential function $V(x)$ to be zero, with $V(x)=\left\|x-x_{f}\right\|_2^2$. The RL policy will generate an action without safety guarantee first as $u_{\text{RL}}(t) = \mu\left(x_t \mid \theta^\mu\right)+\mathcal{N}_t$, where $\mu(\bullet \mid \theta^\mu)$ is a policy parameterized by deep neural networks $\theta^\mu$ and $\mathcal{N}$ is a random process for promoting exploration. Then the barrier function method \cite{ames2016control} ensures that the action $u_{\text{RL}}$ complies with safety constraints\footnote{The state $x$ and $s$, the control/ action $u$ and $a$, terminologies in control theory and reinforcement learning, are used interchangeably here.}. With the presence of the model uncertainty $\omega$ under distributional shift, the CBF of the safety constraint $h(x)\ge 0$, defined in definition\eqref{cbfdef}, can be rewritten using the chain rule as follows: 
\begin{equation}\label{eq:minimaxcbf}
\begin{aligned}
&\sup _{u \in U}\inf_{\substack{\omega\sim p(\omega)\\ p(\omega)\in\mathcal{P} }}\bigg[\frac{\partial h(x)}{\partial x}(\dot{x}(t))+\alpha(h(x))\bigg] \geqslant 0 \\
= &\sup_{u \in U}\inf_{\substack{\omega\sim p(\omega)\\ p(\omega)\in\mathcal{P} }}\bigg[
\underbrace{\frac{\partial h(x)}{\partial x}(f(x)+g(x) u + \omega )+\alpha(h(x))}_{:=H(x,u,\omega)}
\bigg] \geqslant 0 \\
\end{aligned}
\end{equation}
where $\alpha(\bullet)$ is an extended class-$\mathcal{K}$ function. For simplicity, we use a linear function $\kappa(\bullet)$ as the class-$\mathcal{K}$ function.
The infimum tries to evaluate the worst-case safety constraint while the supremum to find a feasible control input such that the safety constraint is still satisfied even with the worst perturbation. While the worst-case criterion is adopted here (possibly with over-conservatism), other ones such as the chance-constrained criterion are compatible with this framework as well.


Using \eqref{eq:minimaxcbf} as the safety shield to solve for the rectified action $u_{\text{R}}$ under the worst-case distribution of $\omega$ leading to the following formulation
\begin{equation}\label{eq:DROminmaxnorm}
\begin{aligned}
&\min _{u_r\in[\underline{u}, \bar{u}]}  ||u_r - u_{\text{RL}}||^2 \\
&\text { s.t. }  \inf_{p(\omega)\in\mathcal{P}} \mathbb{E}_{\omega\sim p(\omega)} \Big\{ H(x,u,\omega)\big|W_d\big(p(\omega), p_0(\omega)\big)\le \rho \Big\} \geqslant 0 \\
\end{aligned}
\end{equation}
which is a bi-level programming. The supreme in \eqref{eq:minimaxcbf} disappears as part of the feasibility of the high-level minimization in \eqref{eq:DROminmaxnorm}. 
However, addressing this bi-level DRO problem can be challenging due to the high computational complexity of its bi-level nature on top of an already challenging distributionally constrained optimization problem at the lower level. 

For the low-level DRO, obtaining the infimum over the expectation of the distributions $p(\omega)$ has been proved difficult. Various techniques, such as convex relaxations, scenario approximations, or sample-based methods, are used to handle the computational challenges associated with DRO. \cite{rockafellar2000optimization, blanchet2019quantifying,yang2020wasserstein}.
That is because the low-level infimum problem requires a search for the worst-case safety violation $H(x,u,\omega)$ within the infinite-dimensional probability space $\mathcal{P}$, making it intractable to solve~\cite{wang2019solving, thekumparampil2019efficient}. As a result, we will demonstrate how we can solve the low-level safety estimation problem in an efficient way based on distributionally robust optimization~\cite{rahimian2019distributionally}. 
Using the Kantorovich duality~\cite{berger2009optimal, wang2022meta}, the infimum in \eqref{eq:DROminmaxnorm} can be further transformed into a more tractable problem 
\begin{equation}\label{eq:DROminmaxnormd}
\begin{aligned}
&\min _{u_r\in[\underline{u}, \bar{u}]} ||u_r - u_{\text{RL}}||^2\\
& \text { s.t. } \mathbb{E}_{\omega_0\sim p_0(\omega)} \inf_{\omega}\Big\{ H(x,u,\omega)\big|d\big(\omega, \omega_0\big)\le \rho_d \Big\} \geqslant 0  \\
\end{aligned}
\end{equation}
where $d(\omega, \omega_0) = \|\omega- \omega_0\|_p^2, p\ge 1$ denotes the the ``cost” for an adversary to perturb $\omega_0$ to $\omega$~\cite{sinha2017certifiable}.
This equivalent dual reformulation~\cite{sinha2017certifiable,wang2022meta} allows for solving the infimum over the noisy state $\omega\in\mathbb{R}^n$, a \textit{parametric finite-dimensional space} instead of over $p(w)$ in the infinite-dimensional probability space $\mathcal{P}$ in~\eqref{eq:DROminmaxnorm}. The radius $\rho_d$ can be evaluated with Monte-Carlo simulation based on the original ambiguity set measured by the Wasserstein metric. However, even with the complexity of DRO alleviated, the problem is still bi-level, and hence in our case, we propose a single-level reformulation to improve tractability without compromising safety.
\subsection{Reduce the bi-level learning to single-level: a differentiable convex programming-based approach}\label{sect: reduce}
To address the complexity of the bi-level optimization problem \eqref{eq:DROminmaxnormd}, we will take advantage of differentiable programming to reduce the problem into a single-level optimization problem. The basic idea is that for general optimization problems, there is a trade-off between optimizing objectives and satisfying constraints. Specifically here, when the dynamics perturbation $\omega$ makes the constraint satisfaction more challenging, then the control input will lean more on constraint satisfaction rather than optimizing the objective. Eventually, the effect of the model uncertainty will be quantified by the objective function values. For example, for a variable value not satisfying the constraints, the objective value tends to go to infinity. As a result, we move the infimum of the constraints to the maximization of the objective function. 
For safe reinforcement learning, the loss function is the negative of the discounted cumulative reward with the safety-proofed actions. We decompose the bi-level problem in \eqref{eq:DROminmaxnormd} into two parts: the safety calibration to get $u_r$ followed by the update of $\omega$ to get the worst case uncertainty. Specifically, for the former, we can start by solving for the optimal action subject to the sample noise under a prior distribution, i.e., $\omega_0\sim p_0(w)$ via the simple CBF-based quadratic program:
\begin{equation}\label{eq:minnorm}
\begin{aligned}
\min _{u_r\in[\underline{u}, \bar{u}]} & ||u_r - u_{\text{RL}}||^2 \\
\text { s.t. } \quad & \mathbb{E}_{\omega_0\sim p_0(\omega)} H(x,u,\omega_0) \geqslant 0  \\
\end{aligned}
\end{equation}
Note that in one-step propagation, multiple samples $\omega_0$'s will be sampled as parameters to evaluate the mean in \eqref{eq:minnorm}.
Then we can propagate our plant using the rectified action $u_r$, and calculate the episodic loss function as
\begin{equation}\label{loss}
\mathcal{L} = -\sum_{k=1}^T \gamma^k {R}(x_k, u_{r,k}).
\end{equation}
By the end of each episode, we can extract a sample of the noise $\omega_{0}$ and the gradient of the loss function with respect to the rectified action $\frac{\partial \mathcal{L}}{\partial u_r}$.
As discussed before, the model uncertainty will try to maximize the loss function, which will be used to find the worst-case perturbation $\omega$ by gradient-based algorithms. Therefore, we can use the chain rule to obtain the gradient of the loss function with respect to the noise $\omega$ as:
\begin{equation}\label{eq:chainerulegradient}
    \frac{\partial \mathcal{L}}{\partial \omega} = \frac{\partial \mathcal{L}}{\partial u_r} \frac{\partial u_r}{\partial \omega}
\end{equation}
To obtain the second term,  differentiable convex programming will be leveraged for solving \eqref{eq:minnorm} and extract the gradient of the rectified action ($u_r$, the variable) with respect to the noise ($\omega$, the parameter) as $\frac{\partial u_r}{\partial \omega}$, evaluated at the samples $\omega_0$'s for trajectory rollout. Note that the gradients flow through the QP \eqref{eq:minnorm} so that the whole pipeline is end-to-end differentiable. 
Since we aim to find the worst-case uncertainty dynamics within the ambiguity set $\mathcal{B} = \{\omega|\mathbb{E}_{\omega_0\sim p_0(w)}d(\omega - \omega_0) \leqslant \rho_d\}$ by maximizing the loss function, projected gradient ascent will be employed to update the $\omega$ as
\begin{equation}\label{eq:GAupdate}
\omega \leftarrow \proj_{\mathcal{B}}\bigg[\omega + \alpha \frac{\partial \mathcal{L}}{\partial \omega}\bigg].
\end{equation}
The projection of $\bar{\omega}$ onto a ball can be done analytically by performing a change of basis to the $l_2$ ball towards the center $\mathbb{E}(\omega_0)$. Then projecting $\bar{\omega}$ onto the closest point within the ball leads to
\begin{equation}\label{eq:proj}
\proj_{\mathcal{B}}(\bar{\omega}) = \mathbb{E}(\omega_0) + \rho_d \frac{\bar{\omega} - \mathbb{E}(\omega_0)}{\max (\rho_d , ||\bar{\omega} - \mathbb{E}(\omega_0)||)}
\end{equation}

\begin{figure}[tbhp]
  \centering
  \vspace{-0.25cm}
  \includegraphics[width=.475\textwidth]{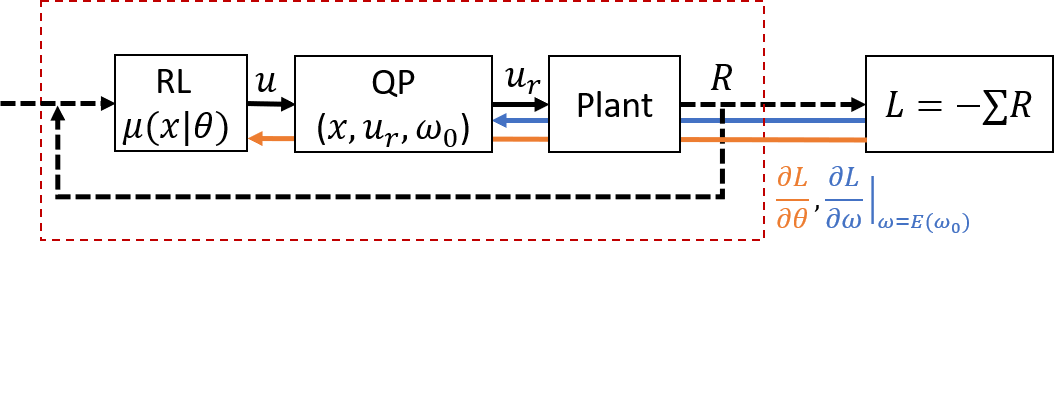}
  \vspace{-1.5cm}
  \caption{Overview of the single-level, end-to-end differentiable convex programming learning-based approach for distributionally safe reinforcement learning using a Wasserstein metric. Black (dashed) lines represent data flow and trajectory rollout, while colored solid lines represent gradient back-propagation for updating both policies and worst-case noise.}
  \label{fig: workflow}
  \vspace{-.5cm}
\end{figure}

Fig.\ref{fig: workflow} illustrates the workflow of the proposed approach.
Algorithm\ref{DSRL} summarizes the overall distributional safe RL framework with DDPG~\cite{lillicrap2015continuous} and the learnable worst-case noise $\omega$ under a distributional shift. In summary, we transformed bi-level programming for distributional safety into a sequential problem: a convex quadratic program followed by a projected gradient ascent learning module for the worst-case noise. Both parts are computationally cheap and differentiable convex programming is employed for end-to-end differentiability to update both policy parameters with constraints and worst-case uncertainty.

\begin{algorithm}[h]
\caption{Distributionally Safe Reinforcement Learning}\label{DSRL}
\begin{algorithmic}[1]
\State \textbf{Require:} Environment setting, learning rates $\alpha, \beta$, discount factor $\gamma$, target network update rate $\tau$, and samples $\omega_0 \sim p_0(\omega)$.
\State Initialize critic network  $Q\left(s, a \mid \theta^Q\right)$, actor $\mu\left(s \mid \theta^\mu\right)$ with weights $\theta^Q$ and $\theta^\mu$.
\State Initialize target network $Q^{\prime}$ and $\mu^{\prime}$ with weights $\theta^{Q^{\prime}} \leftarrow \theta^Q, \theta^{\mu^{\prime}} \leftarrow \theta^\mu$ .
\State Initialize replay buffer $\mathcal{D}$.
\State Initialize a random $\omega$.
\For{episode $=1,\ldots,M$}
\State Initialize a random noise $\omega_0 \sim p_0(\omega)$ from samples .
\State Initialize a random process $\mathcal{N}$ for action exploration.
\State Receive initial observation state $s_1$ .
\For{${t}=1,\ldots,T$}
\State \parbox[t]{200pt}{Select action $a_t=\mu\left(s_t \mid \theta^\mu\right)+\mathcal{N}_t$ according to the current policy and exploration noise. \strut}
\State \parbox[t]{200pt}{ \underline{Get rectified action $a_{t_R}$ via \eqref{eq:minnorm} and samples $\omega_0$.}}
\State \parbox[t]{200pt}{Execute action $a_{t_R}$ and observe reward $R_t$ and new state $s_{t+1}$.\strut}
\State Store transition \underline{$\left(s_t, a_t, a_{t_R}, R_t, s_{t+1}, \omega_0\right)$} in $\mathcal{D}$.
\State \parbox[t]{200pt}{Sample a random mini-batch of ${N}$ transitions $\left(s_t, a_t, a_{t_R}, R_t, s_{t+1}, \omega_0\right)$ from $\mathcal{D}$.\strut} 
\State \parbox[t]{200pt}{Update critic using learning rate $\beta$. \strut}  
\State \parbox[t]{200pt}{Update the actor $\theta^\mu$ using the gradient ascent with the sampled gradient of the return in \eqref{loss}. \strut}  
\State $\theta^\mu \leftarrow \theta^\mu + \alpha \nabla_{\theta^\mu}J(\theta).$
\State Update the target networks with rate $\tau$ .
\State $\theta^{\prime} \leftarrow \tau \theta+(1-\tau) \theta^\prime$ . 
\EndFor
\State \parbox[t]{200pt}{Sample noise $\omega_0$ from all transitions $\left(s_t, a_t, a_{t_R}, R_t, s_{t+1}, \omega_0\right)$ in $\mathcal{D}$ and evaluate the loss $\mathcal{L}$ and its gradient $\frac{\partial \mathcal{L}}{\partial \omega}$. \strut}  
\State \parbox[t]{200pt}{Update $\omega$ using the projected gradient ascent \\ 
\underline{$\omega \leftarrow \proj_{\mathcal{B}}\big[\omega + \alpha \frac{\partial \mathcal{L}}{\partial \omega}\big]$} with the gradeient evaluation in \eqref{eq:chainerulegradient} and the projection in \eqref{eq:proj}. \strut} 
\EndFor 
\State \textbf{Return:} $\theta^\mu, \theta^Q, \omega$.
\end{algorithmic}
\end{algorithm}

\section{SIMULATIONS AND RESULTS}\label{sect: numerical}
In this section, we assess the performance of the suggested safe reinforcement learning approach using two cases of Dubin's car, a first-order and second-order system, and a simplified 3D quadcopter. Our goal is to compare the performance of the deterministically learned policy in the presence of dynamics uncertainty, with the distributionally safe learned policy. The comparison is carried out under the condition that all other settings and parameters are kept identical to ensure a consistent assessment. By carrying out this comparison, we aim to show the effectiveness of the distributionally safe approach in dealing with uncertain environments.

\subsubsection{\textbf{First-Order Dubins Car}}
The first simulation is carried out using the first-order Dubins car environment with the following kinematics\eqref{dubinskinematics}.
\begin{equation}\label{dubinskinematics}
\left(\begin{array}{c}
\dot{x} \\
\dot{y} \\
\dot{\theta}
\end{array}\right)=\left[\begin{array}{ccc}
\cos \theta & -\sin \theta & 0 \\
\sin \theta & \cos \theta & 0 \\
0 & 0 & 1
\end{array}\right]\left(\begin{array}{c}
u_x \\
u_y \\
u_\theta
\end{array}\right),
\end{equation}
where $u_x, u_y, u_\theta$ are the input velocity along the $x$ axis, the sideways input velocity, and the angular input velocity, respectively. To reach its final destination $x_f$ from an initial state $x_o$, we use a reward that penalizes the squared distance between the car and the goal state multiplied by a coefficient as $d\left\|x-x_{f}\right\|_2^2$, and penalizes every time step by a constant $s$ for minimum time goal-reaching. Hence, the reward is defined as $R = -d\left\|x-x_{f}\right\|_2^2 - s$, with $d>0$ and $s\ge 0$.

\begin{figure}[thpb]
  \centering
  \vspace{-0.4cm}
  \includegraphics[scale=0.4]{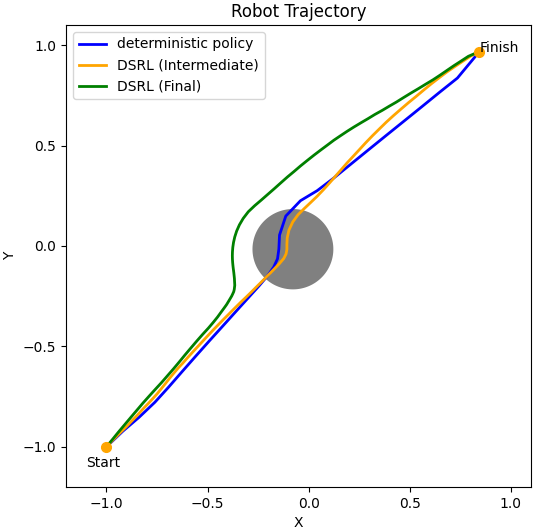}
  \vspace{-.45cm}
  \caption{First-order Dubin's car trajectories with the presence of model uncertainty for deterministic policy, the DSRL policies before and after convergence.}
  \label{1traj}
  \vspace{-.4cm}
\end{figure}

Fig. \ref{1traj} presents a comparison between trajectories generated by a deterministically learned policy, and distributionally safe policies before and after training converges.

 We can see that although the deterministic policy is trying to avoid the obstacle by swerving in the same shape as the obstacle, it still violated the safety constraint because it could not take into account the shift/ noise present in the dynamics of the system. The DSRL policy adapted to the noise after several episodes of training and started avoiding the obstacle while keeping a very safe margin.

 \begin{figure}[t] 
\minipage{0.24\textwidth} \includegraphics[width=0.99\linewidth, height=0.99\linewidth]{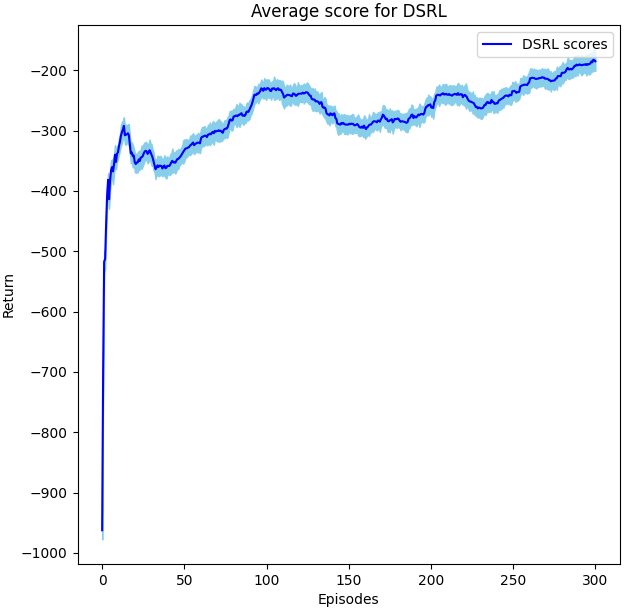} \endminipage\hfill
\minipage{0.24\textwidth} \includegraphics[width=0.99\linewidth, height=0.99\linewidth]{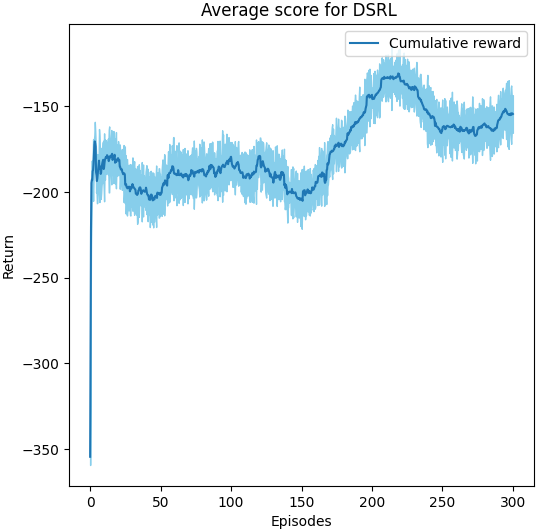} \endminipage\hfill
\vspace{-0.2cm}
\caption{DSRL returns over the learning period for 3 random seeds for the Dubin's car: Left: first-order; Right: second order.
}
\vspace{-.6cm}
\label{fig:dubins}
\end{figure}


 In Fig. \ref{fig:dubins} (left), we can see that although the learning of the DSRL policy is fluctuating between episodes, the general trend of the cumulative reward increases over the total number of episodes, demonstrating a promising learning curve.

\subsubsection{\textbf{Second-Order Dubins Car}}
\begin{figure}[thpb]
  \centering
  \vspace{-0.4cm}
  \includegraphics[scale=0.4]{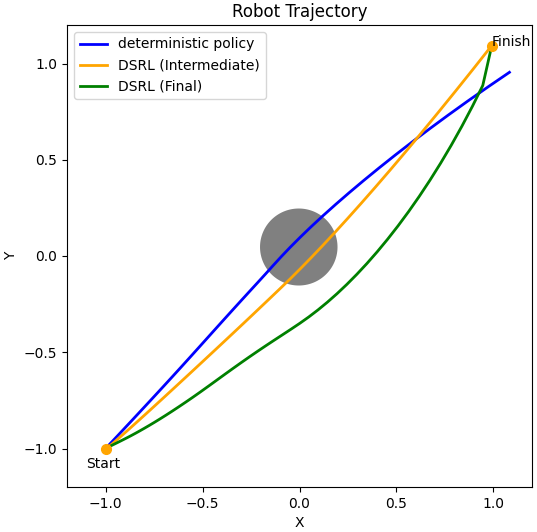}
  \vspace{-0.45cm}
  \caption{Second-order Dubin's car trajectories with the presence of model uncertainty for deterministic policy, the DSRL policies before and after convergence.}
  \label{2traj}
  \vspace{-.75cm}
\end{figure}

To demonstrate our approach on a higher-order system, we use the second-order Dubin's car with the following dynamics:
\begin{equation}\label{2dubinskinematics}
\left(\begin{array}{c}
\ddot{x} \\
\ddot{y} \\
\ddot{\theta}
\end{array}\right)=\left[\begin{array}{ccc}
\cos \theta & -\sin \theta & 0 \\
\sin \theta & \cos \theta & 0 \\
0 & 0 & 1
\end{array}\right]\left(\begin{array}{c}
u_x \\
u_y \\
\tau_c
\end{array}\right),
\end{equation}
We also have the following adjusted reward function to penalize the velocities at the final destination for learning to brake as well: $R = -d\left\|x-x_{f}\right\|_2^2 -b\left\|v-v_{f}\right\|_2^2 - s$.
Fig. \ref{2traj} and Fig.\ref{fig:dubins}(right) illustrate the trajectories of the benchmark and two episodes from the learning episodes, and the cumulative rewards. In the second-order case, we can see that the benchmark misses the finish line by a slight margin due to the influence of noise. All other behaviors are similar to the first-order case.


\subsubsection{\textbf{Quadcopter}}

To inspect the performance of the DSRL on a system with more complex dynamics and safety constraints, we design high-level controllers for a heading-locked quadcopter environment with the following dynamics

\begin{equation}\label{Quadkinematics}
\left(\begin{array}{c}
\dot{x} \\
\dot{y} \\
\dot{z}
\end{array}\right)=\left(\begin{array}{c}
T\sin{\theta} \\
T\cos{\theta}\sin{\phi} \\
T\cos{\theta}\cos{\phi} - g
\end{array}\right)
\end{equation}
where $x, y, z$ are the inertial displacements of the quadcopter and $\theta, \phi, T$ are the pitch, roll, and total thrust, respectively. While training, the quadcopter is penalized by the distance between the initial and terminal positions: $ R = -\left\|x-x_{f}\right\|_2^2$.
In this section, we simulate a landing scenario of the quadcopter within a prescribed glide slope defined by a cone with a half angle $\delta$. The safety constraint can be written concisely as:
\begin{equation}
r_{I}^TM_{g s} r_{I} \geqslant 0
\end{equation}
where $M_{gs}=[[-\cot(\delta_{gs})^{2} , 0 , 0]^{T}, [0 , -\cot(\delta_{gs})^{2} , 0]^{T}, [0 , 0 , 1]^{T}]$ and $r_I=[x,y,z]$ is the inertial position of the quadcopter.

\begin{figure}[thpb]
  \centering
  \vspace{-.2cm}
  \includegraphics[scale=0.5]{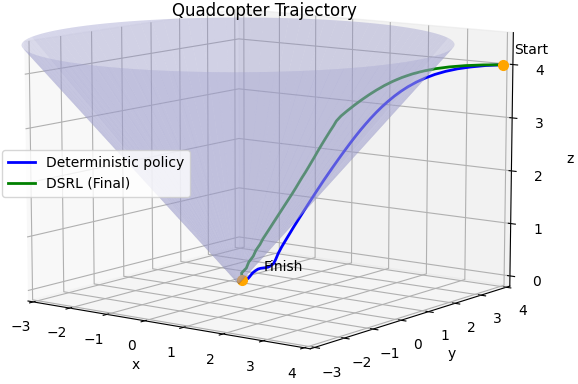}
  \vspace{-.4cm}
  \caption{Quadcopter trajectories with the presence of model uncertainty for deterministic policy and the DSRL policy after convergence. }
  \vspace{-.4cm}
  \label{quadtrajectory}
  \vspace{-.0cm}
\end{figure}


Fig.\ref{quadtrajectory} presents the trajectories generated by the DSRL at the start and end of the learning, and the one generated by the deterministic policy subject to dynamic noise. The learned $\omega$ outperforms the deterministic policy, as it always keeps a safe margin from the constraint, hence, maintaining safety throughout the whole trajectory.

\section{CONCLUSIONS}\label{sect: conc}
This work addresses the challenges of tractable safe reinforcement learning under distributional shift. We employ differentiable convex programming along with distributinoally robust optimization to enable the safe learning over unpredictable probability distributions of the model's uncertainty in a tractable manner. We evaluate our approach on first and second-order Dubin's car and a simplified quad-copter model and compare our results with the deterministic policies' performance, where our approach outperforms the benchmark policies regarding safety guarantee under model uncertainty.






\bibliography{root.bib}

\begin{thebibliography}{10}
\providecommand{\url}[1]{#1}
\csname url@rmstyle\endcsname
\providecommand{\newblock}{\relax}
\providecommand{\bibinfo}[2]{#2}
\providecommand\BIBentrySTDinterwordspacing{\spaceskip=0pt\relax}
\providecommand\BIBentryALTinterwordstretchfactor{4}
\providecommand\BIBentryALTinterwordspacing{\spaceskip=\fontdimen2\font plus
\BIBentryALTinterwordstretchfactor\fontdimen3\font minus \fontdimen4\font\relax}
\providecommand\BIBforeignlanguage[2]{{%
\expandafter\ifx\csname l@#1\endcsname\relax
\typeout{** WARNING: IEEEtran.bst: No hyphenation pattern has been}%
\typeout{** loaded for the language `#1'. Using the pattern for}%
\typeout{** the default language instead.}%
\else
\language=\csname l@#1\endcsname
\fi
#2}}

\bibitem{li2019robust}
S.~Li, Y.~Wu, X.~Cui, H.~Dong, F.~Fang, and S.~Russell, ``Robust multi-agent reinforcement learning via minimax deep deterministic policy gradient,'' in \emph{Proceedings of the AAAI Conference on Artificial Intelligence}, vol.~33, no.~01, 2019, pp. 4213--4220.

\bibitem{zhang2019policy}
K.~Zhang, Z.~Yang, and T.~Basar, ``Policy optimization provably converges to nash equilibria in zero-sum linear quadratic games,'' \emph{Advances in Neural Information Processing Systems}, vol.~32, 2019.

\bibitem{sun2021romax}
C.~Sun, D.-K. Kim, and J.~P. How, ``Romax: Certifiably robust deep multiagent reinforcement learning via convex relaxation,'' \emph{arXiv preprint arXiv:2109.06795}, 2021.

\bibitem{yang2022stackelberg}
B.~Yang, L.~Zheng, L.~J. Ratliff, B.~Boots, and J.~R. Smith, ``Stackelberg maddpg: Learning emergent behaviors via information asymmetry in competitive games,'' 2022.

\bibitem{zhou2021decentralized}
Z.~Zhou and H.~Xu, ``Decentralized adaptive optimal tracking control for massive autonomous vehicle systems with heterogeneous dynamics: A stackelberg game,'' \emph{IEEE Transactions on Neural Networks and Learning Systems}, vol.~32, no.~12, pp. 5654--5663, 2021.

\bibitem{lauffer2022no}
N.~Lauffer, M.~Ghasemi, A.~Hashemi, Y.~Savas, and U.~Topcu, ``No-regret learning in dynamic stackelberg games,'' \emph{arXiv preprint arXiv:2202.04786}, 2022.

\bibitem{bai2021sample}
Y.~Bai, C.~Jin, H.~Wang, and C.~Xiong, ``Sample-efficient learning of stackelberg equilibria in general-sum games,'' \emph{Advances in Neural Information Processing Systems}, vol.~34, 2021.

\bibitem{jin2021safe}
W.~Jin, S.~Mou, and G.~Pappas, ``Safe pontryagin differentiable programming,'' \emph{Advances in Neural Information Processing Systems}, vol.~34, 2021.

\bibitem{liu2020ipo}
Y.~Liu, J.~Ding, and X.~Liu, ``Ipo: Interior-point policy optimization under constraints,'' in \emph{Proceedings of the AAAI Conference on Artificial Intelligence}, vol.~34, no.~04, 2020, pp. 4940--4947.

\bibitem{achiam2017constrained}
J.~Achiam, D.~Held, A.~Tamar, and P.~Abbeel, ``Constrained policy optimization,'' in \emph{Proceedings of the 34th International Conference on Machine Learning-Volume 70}.\hskip 1em plus 0.5em minus 0.4em\relax JMLR. org, 2017, pp. 22--31.

\bibitem{bertsekas2015parallel}
D.~Bertsekas and J.~Tsitsiklis, \emph{Parallel and distributed computation: numerical methods}.\hskip 1em plus 0.5em minus 0.4em\relax Athena Scientific, 2015.

\bibitem{geibel2005risk}
P.~Geibel and F.~Wysotzki, ``Risk-sensitive reinforcement learning applied to control under constraints,'' \emph{Journal of Artificial Intelligence Research}, vol.~24, pp. 81--108, 2005.

\bibitem{chow2017risk}
Y.~Chow, M.~Ghavamzadeh, L.~Janson, and M.~Pavone, ``Risk-constrained reinforcement learning with percentile risk criteria,'' \emph{The Journal of Machine Learning Research}, vol.~18, no.~1, pp. 6070--6120, 2017.

\bibitem{chow2018lyapunov}
Y.~Chow, O.~Nachum, E.~Duenez-Guzman, and M.~Ghavamzadeh, ``A lyapunov-based approach to safe reinforcement learning,'' in \emph{Advances in neural information processing systems}, 2018, pp. 8092--8101.

\bibitem{stooke2020responsive}
A.~Stooke, J.~Achiam, and P.~Abbeel, ``Responsive safety in reinforcement learning by pid lagrangian methods,'' \emph{arXiv preprint arXiv:2007.03964}, 2020.

\bibitem{amos2017optnet}
B.~Amos and J.~Z. Kolter, ``Optnet: Differentiable optimization as a layer in neural networks,'' \emph{arXiv preprint arXiv:1703.00443}, 2017.

\bibitem{zhang2020robust1}
H.~Zhang, H.~Chen, C.~Xiao, B.~Li, M.~Liu, D.~Boning, and C.-J. Hsieh, ``Robust deep reinforcement learning against adversarial perturbations on state observations,'' \emph{Advances in Neural Information Processing Systems}, vol.~33, pp. 21\,024--21\,037, 2020.

\bibitem{zhang2020robust}
K.~Zhang, T.~Sun, Y.~Tao, S.~Genc, S.~Mallya, and T.~Basar, ``Robust multi-agent reinforcement learning with model uncertainty,'' \emph{Advances in Neural Information Processing Systems}, vol.~33, pp. 10\,571--10\,583, 2020.

\bibitem{brunke2021safe}
L.~Brunke, M.~Greeff, A.~W. Hall, c.~Yuan, S.~Zhou, J.~Panerati, and A.~P. Schoellig, ``Safe learning in robotics: From learning-based control to safe reinforcement learning,'' \emph{Annual Review of Control, Robotics, and Autonomous Systems}, vol.~5, 2021.

\bibitem{singh2021reinforcement}
B.~Singh, R.~Kumar, and V.~P. Singh, ``Reinforcement learning in robotic applications: a comprehensive survey,'' \emph{Artificial Intelligence Review}, pp. 1--46, 2021.

\bibitem{garcia2015comprehensive}
J.~Garc{\i}a and F.~Fern{\'a}ndez, ``A comprehensive survey on safe reinforcement learning,'' \emph{Journal of Machine Learning Research}, vol.~16, no.~1, pp. 1437--1480, 2015.

\bibitem{moos2022robust}
J.~Moos, K.~Hansel, H.~Abdulsamad, S.~Stark, D.~Clever, and J.~Peters, ``Robust reinforcement learning: A review of foundations and recent advances,'' \emph{Machine Learning and Knowledge Extraction}, vol.~4, no.~1, pp. 276--315, 2022.

\bibitem{eghbal2021learning}
H.~Eghbal-zadeh, F.~Henkel, and G.~Widmer, ``Learning to infer unseen contexts in causal contextual reinforcement learning,'' in \emph{Self-Supervision for Reinforcement Learning Workshop-ICLR 2021}, 2021.

\bibitem{rakelly2019efficient}
K.~Rakelly, A.~Zhou, C.~Finn, S.~Levine, and D.~Quillen, ``Efficient off-policy meta-reinforcement learning via probabilistic context variables,'' in \emph{International conference on machine learning}.\hskip 1em plus 0.5em minus 0.4em\relax PMLR, 2019, pp. 5331--5340.

\bibitem{choi2020reinforcement}
J.~Choi, F.~Casta{\~n}eda, C.~J. Tomlin, and K.~Sreenath, ``Reinforcement learning for safety-critical control under model uncertainty, using control lyapunov functions and control barrier functions,'' \emph{arXiv preprint arXiv:2004.07584}, 2020.

\bibitem{zheng2021safe}
L.~Zheng, Y.~Shi, L.~J. Ratliff, and B.~Zhang, ``Safe reinforcement learning of control-affine systems with vertex networks,'' in \emph{Learning for Dynamics and Control}.\hskip 1em plus 0.5em minus 0.4em\relax PMLR, 2021, pp. 336--347.

\bibitem{cheng2019end}
R.~Cheng, G.~Orosz, R.~M. Murray, and J.~W. Burdick, ``End-to-end safe reinforcement learning through barrier functions for safety-critical continuous control tasks,'' in \emph{Proceedings of the AAAI Conference on Artificial Intelligence}, vol.~33, 2019, pp. 3387--3395.

\bibitem{ames2016control}
A.~D. Ames, X.~Xu, J.~W. Grizzle, and P.~Tabuada, ``Control barrier function based quadratic programs for safety critical systems,'' \emph{IEEE Transactions on Automatic Control}, vol.~62, no.~8, pp. 3861--3876, 2016.

\bibitem{berkenkamp2017safe}
F.~Berkenkamp, M.~Turchetta, A.~Schoellig, and A.~Krause, ``Safe model-based reinforcement learning with stability guarantees,'' in \emph{Advances in neural information processing systems}, 2017, pp. 908--918.

\bibitem{sun2021fisar}
C.~Sun, D.-K. Kim, and J.~P. How, ``Fisar: Forward invariant safe reinforcement learning with a deep neural network-based optimizer,'' in \emph{2021 IEEE International Conference on Robotics and Automation (ICRA)}.\hskip 1em plus 0.5em minus 0.4em\relax IEEE, 2021, pp. 10\,617--10\,624.

\bibitem{shi2021meta}
G.~Shi, K.~Azizzadenesheli, M.~O'Connell, S.-J. Chung, and Y.~Yue, ``Meta-adaptive nonlinear control: Theory and algorithms,'' \emph{Advances in Neural Information Processing Systems}, vol.~34, pp. 10\,013--10\,025, 2021.

\bibitem{o2022neural}
M.~O’Connell, G.~Shi, X.~Shi, K.~Azizzadenesheli, A.~Anandkumar, Y.~Yue, and S.-J. Chung, ``Neural-fly enables rapid learning for agile flight in strong winds,'' \emph{Science Robotics}, vol.~7, no.~66, p. eabm6597, 2022.

\bibitem{van2015distributionally}
B.~P. Van~Parys, D.~Kuhn, P.~J. Goulart, and M.~Morari, ``Distributionally robust control of constrained stochastic systems,'' \emph{IEEE Transactions on Automatic Control}, vol.~61, no.~2, pp. 430--442, 2015.

\bibitem{yang2020wasserstein}
I.~Yang, ``Wasserstein distributionally robust stochastic control: A data-driven approach,'' \emph{IEEE Transactions on Automatic Control}, vol.~66, no.~8, pp. 3863--3870, 2020.

\bibitem{hakobyan2021wasserstein}
A.~Hakobyan and I.~Yang, ``Wasserstein distributionally robust motion control for collision avoidance using conditional value-at-risk,'' \emph{IEEE Transactions on Robotics}, vol.~38, no.~2, pp. 939--957, 2021.

\bibitem{bahari2022safe}
A.~Bahari~Kordabad, R.~Wisniewski, and S.~Gros, ``Safe reinforcement learning using wasserstein distributionally robust mpc and chance constraint,'' 2022.

\bibitem{coulson2021distributionally}
J.~Coulson, J.~Lygeros, and F.~D{\"o}rfler, ``Distributionally robust chance constrained data-enabled predictive control,'' \emph{IEEE Transactions on Automatic Control}, vol.~67, no.~7, pp. 3289--3304, 2021.

\bibitem{coppens2021data}
P.~Coppens and P.~Patrinos, ``Data-driven distributionally robust mpc for constrained stochastic systems,'' \emph{IEEE Control Systems Letters}, vol.~6, pp. 1274--1279, 2021.

\bibitem{ren2022distributionally}
A.~Z. Ren and A.~Majumdar, ``Distributionally robust policy learning via adversarial environment generation,'' \emph{IEEE Robotics and Automation Letters}, vol.~7, no.~2, pp. 1379--1386, 2022.

\bibitem{morrison2020egad}
D.~Morrison, P.~Corke, and J.~Leitner, ``Egad! an evolved grasping analysis dataset for diversity and reproducibility in robotic manipulation,'' \emph{IEEE Robotics and Automation Letters}, vol.~5, no.~3, pp. 4368--4375, 2020.

\bibitem{wang2019adversarial}
D.~Wang, D.~Tseng, P.~Li, Y.~Jiang, M.~Guo, M.~Danielczuk, J.~Mahler, J.~Ichnowski, and K.~Goldberg, ``Adversarial grasp objects,'' in \emph{2019 IEEE 15th International Conference on Automation Science and Engineering (CASE)}.\hskip 1em plus 0.5em minus 0.4em\relax IEEE, 2019, pp. 241--248.

\bibitem{xu2022group}
M.~Xu, P.~Huang, Y.~Niu, V.~Kumar, J.~Qiu, C.~Fang, K.-H. Lee, X.~Qi, H.~Lam, B.~Li, \emph{et~al.}, ``Group distributionally robust reinforcement learning with hierarchical latent variables,'' \emph{arXiv preprint arXiv:2210.12262}, 2022.

\bibitem{kallus2022doubly}
N.~Kallus, X.~Mao, K.~Wang, and Z.~Zhou, ``Doubly robust distributionally robust off-policy evaluation and learning,'' in \emph{International Conference on Machine Learning}.\hskip 1em plus 0.5em minus 0.4em\relax PMLR, 2022, pp. 10\,598--10\,632.

\bibitem{shi2022distributionally}
L.~Shi and Y.~Chi, ``Distributionally robust model-based offline reinforcement learning with near-optimal sample complexity,'' \emph{arXiv preprint arXiv:2208.05767}, 2022.

\bibitem{sinha2017certifiable}
A.~Sinha, H.~Namkoong, and J.~Duchi, ``Certifiable distributional robustness with principled adversarial training,'' \emph{arXiv preprint arXiv:1710.10571}, vol.~2, 2017.

\bibitem{blanchet2019quantifying}
J.~Blanchet and K.~Murthy, ``Quantifying distributional model risk via optimal transport,'' \emph{Mathematics of Operations Research}, vol.~44, no.~2, pp. 565--600, 2019.

\bibitem{rahimian2019distributionally}
H.~Rahimian and S.~Mehrotra, ``Distributionally robust optimization: A review,'' \emph{arXiv preprint arXiv:1908.05659}, 2019.

\bibitem{xiaohigh}
W.~Xiao and C.~Belta, ``High-order control barrier functions,'' \emph{IEEE Transactions on Automatic Control}, vol.~67, no.~7, pp. 3655--3662, 2021.

\bibitem{diffcvxoptlayer}
A.~Agrawal, B.~Amos, S.~Barratt, S.~Boyd, S.~Diamond, and J.~Z. Kolter, ``Differentiable convex optimization layers,'' \emph{Advances in neural information processing systems}, vol.~32, 2019.

\bibitem{diffconeprog}
A.~Agrawal, S.~Barratt, S.~Boyd, E.~Busseti, and W.~M. Moursi, ``Differentiating through a cone program,'' \emph{arXiv preprint arXiv:1904.09043}, 2019.

\bibitem{chriat2023optimality}
A.~E. Chriat and C.~Sun, ``On the optimality, stability, and feasibility of control barrier functions: An adaptive learning-based approach,'' \emph{arXiv preprint arXiv:2305.03608}, 2023.

\bibitem{dacs2023robust}
E.~Da{\c{s}}, S.~X. Wei, and J.~W. Burdick, ``Robust control barrier functions with uncertainty estimation,'' \emph{arXiv preprint arXiv:2304.08538}, 2023.

\bibitem{chriat2023CVar}
A.~E. Chriat and C.~Sun, ``Wasserstein distributionally robust control barrier function using conditional value-at-risk with differentiable convex programming,'' \emph{Accepted to AIAA SciTech 2024}, 2024.

\bibitem{givens1984class}
C.~R. Givens and R.~M. Shortt, ``A class of wasserstein metrics for probability distributions.'' \emph{Michigan Mathematical Journal}, vol.~31, no.~2, pp. 231--240, 1984.

\bibitem{joyce2011kullback}
J.~M. Joyce, ``Kullback-leibler divergence,'' in \emph{International encyclopedia of statistical science}.\hskip 1em plus 0.5em minus 0.4em\relax Springer, 2011, pp. 720--722.

\bibitem{rockafellar2000optimization}
R.~T. Rockafellar, S.~Uryasev, \emph{et~al.}, ``Optimization of conditional value-at-risk,'' \emph{Journal of risk}, vol.~2, pp. 21--42, 2000.

\bibitem{wang2019solving}
Y.~Wang, G.~Zhang, and J.~Ba, ``On solving minimax optimization locally: A follow-the-ridge approach,'' \emph{arXiv preprint arXiv:1910.07512}, 2019.

\bibitem{thekumparampil2019efficient}
K.~K. Thekumparampil, P.~Jain, P.~Netrapalli, and S.~Oh, ``Efficient algorithms for smooth minimax optimization,'' \emph{Advances in Neural Information Processing Systems}, vol.~32, 2019.

\bibitem{berger2009optimal}
M.~Berger, B.~Eckmann, P.~Harpe, F.~Hirzebruch, N.~Hitchin, L.~H{\"o}rmander, A.~Kupiainen, G.~Lebeau, M.~Ratner, D.~Serre, \emph{et~al.}, \emph{Optimal transport: old and new}.\hskip 1em plus 0.5em minus 0.4em\relax Springer, 2009.

\bibitem{wang2022meta}
Z.~Wang, X.~Wang, L.~Shen, Q.~Suo, K.~Song, D.~Yu, Y.~Shen, and M.~Gao, ``Meta-learning without data via wasserstein distributionally-robust model fusion,'' in \emph{Uncertainty in Artificial Intelligence}.\hskip 1em plus 0.5em minus 0.4em\relax PMLR, 2022, pp. 2045--2055.

\bibitem{lillicrap2015continuous}
T.~P. Lillicrap, J.~J. Hunt, A.~Pritzel, N.~Heess, T.~Erez, Y.~Tassa, D.~Silver, and D.~Wierstra, ``Continuous control with deep reinforcement learning,'' \emph{arXiv preprint arXiv:1509.02971}, 2015.

\end{thebibliography}

\bibliographystyle{IEEEtran}
\end{document}